\documentclass[10pt,twocolumn,letterpaper]{article}

\usepackage{cvpr}              

\usepackage[dvipsnames]{xcolor}

\usepackage{times}  
\usepackage{helvet}  
\usepackage{courier}  
\usepackage[hyphens]{url}  
\usepackage{graphicx} 
\usepackage{natbib}  
\usepackage{caption} 
\usepackage{amssymb}
\usepackage{pifont}
\usepackage{colortbl}
\usepackage{amsmath}
\usepackage{multirow}

\definecolor{shadecolor}{rgb}{0.92, 0.92, 0.92}
\definecolor{gtgray}{gray}{0.97}
\definecolor{mygray}{gray}{.88}

\definecolor{gray1}{gray}{.90}
\definecolor{gray2}{gray}{.92}
\definecolor{gray3}{gray}{.94}

\makeatletter
\def\hlinew#1{%
  \noalign{\ifnum0=`}\fi\hrule \@height #1 \futurelet
   \reserved@a\@xhline}
\makeatother

\definecolor{cvprblue}{rgb}{0.21,0.49,0.74}
\usepackage[pagebackref,breaklinks,colorlinks,citecolor=cvprblue]{hyperref}

\def\paperID{212} 
\def\confName{3DV\xspace}
\def\confYear{2025\xspace}

\title{GaussianStyle: Gaussian Head Avatar via StyleGAN}

\author{
Pinxin Liu$^{1}$ $\quad$ Luchuan Song$^{1}$ $\quad$ Daoan Zhang$^{1}$ $\quad$ Yunlong Tang$^{1}$ $\quad$ \\
Hang Hua$^{1}$ $\quad$ Huaijin Tu$^{2}$ \quad Jiebo Luo$^{1}$ $\quad $ Chenliang Xu$^{1}$\\
$^{1}$University of Rochester $\quad$ $^{2}$Georgia Institute of Technology\\
{\tt\small \{pliu23, lsong11, daoan.zhang\}@ur.rochester.edu} \\
{\tt\small \{yunlong.tang, hhua2@cs., jluo@cs., chenliang.xu@\}rochester.edu}\\
{\tt\small htu35@gatech.edu}\\
}

\begin{document}

\twocolumn[{%
\renewcommand\twocolumn[1][]{#1}%
\maketitle
\begin{center}
    \centering
    \vspace{-0.7cm}
    \captionsetup{type=figure}
    \includegraphics[width=1.0\textwidth]{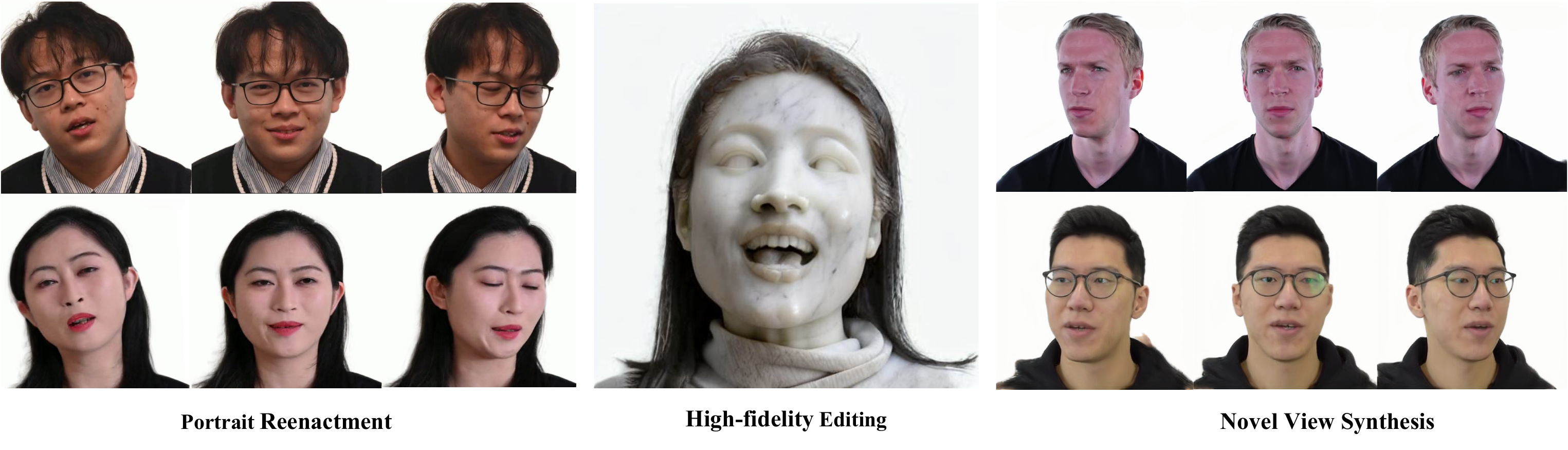}
    \vspace{-0.7cm}
    \captionof{figure}{We present \textbf{GaussianStyle}, a novel method designed for high-fidelity volumetric avatar reconstruction from a short monocular video. Our pipeline can be utilized for portrait reenactment, high-fidelity editing, and novel view synthesis.}
\label{Teaser}
\end{center}%
}]

\begin{abstract}
Existing methods like Neural Radiation Fields (NeRF) and 3D Gaussian Splatting (3DGS) have made significant strides in facial attribute control such as facial animation and components editing, yet they struggle with fine-grained representation and scalability in dynamic head modeling. To address these limitations, we propose GaussianStyle, a novel framework that integrates the volumetric strengths of 3DGS with the powerful implicit representation of StyleGAN. The GaussianStyle preserves structural information, such as expressions and poses, using Gaussian points, while projecting the implicit volumetric representation into StyleGAN to capture high-frequency details and mitigate the over-smoothing commonly observed in neural texture rendering. Experimental outcomes indicate that our method achieves state-of-the-art performance in reenactment, novel view synthesis, and animation.
\end{abstract}    
\section{Introduction}
\label{sec:intro}

Learning head avatars from a given monocular video has become popular in recent years. It aims to achieve diversity control in terms of facial expression and head pose. Many works incorporate NeRF~\cite{INSTA:CVPR2023, song2024tri2plane, zheng2022avatar, Gao2022nerfblendshape, gafni2021dynamic} and 3DGS~\cite{wang2024gaussianhead, xu2023gaussianheadavatar} into head avatar training via tracked parametric facial template. Generally, those methods are maintain a relatively canonical feature space by the implicit topology (for NeRF) or explicit topology (for 3DGS), and enable the queried voxel or Gaussian points to learn the neural texture features from the movement of head in video. 

Though this strategy improve the movement stability, it overlooks a critical challenge inherent within dynamic 3D head modeling: the assumption that a fixed 3D coordinate in the canonical space will always correspond to the same facial region throughout the entire sequence. In reality, as the head motion and dynamic expressions, the relative positions of facial features will shift significantly. For example, a point that initially corresponds to the corner of the mouth in a canonical expression will shift toward the cheek when smiles or lip motions. This movement causes the fixed coordinate canonical template to misalign with the actual facial regions it is supposed to observe.

This limitation manifests as over-smoothing in dynamic head and facial movement rendering scenarios. Since the fixed 3D coordinates do not accurately track the evolving geometry of the face, the model tends to produce averaged or blurred features rather than sharp and precise details. This over-smoothing effect is particularly pronounced in areas where there is a high degree of motion or expression variation, resulting in a loss of the fine-grained details necessary for realistic and expressive head avatars. This problem is further exacerbated during the cross-reenactment scenarios when the motion is conditioned on novel expression, pose, or camera perspectives.

Drawing inspiration from the deferred neural rendering~\cite{thies2019deferred} (DNR), which first samples the noised UV space features and then leverages the neural network to translate the texture space to pixel space, we believe this coarse-to-fine strategy has the potential to address the over-smoothing issue. However, this UV-projected texture (or called neural texture) is difficult to extend to 3D space, leading to limitations in novel-view synthesis and flexible control. 

To address these challenges, we propose \textbf{GaussianStyle}, a novel framework that integrates dynamic neural rendering with 3DGS. By leveraging the powerful implicit representation of StyleGAN, GaussianStyle improves the fine-grained texture quality based on the volumetric representation provided by 3DGS. Specifically, we first propose a more robust dynamic Gaussian representation. Inspired by the Triplane-Gaussian~\cite{zou2023triplane}, we construct a temporal-aware Tri-plane as an implicit and low-dimensional Gaussian representation. This design allows for more effective 4D Gaussian modeling by leveraging cross-attention to learn the correspondence between Gaussian points and motion-control parameters. To extend StyleGAN with 3DGS features, we introduce a multi-view PTI initialization that minimizes disruptions to pre-trained StyleGAN parameters while personalizing the rendering for the target avatar. Additionally, we propose an optimal method for projecting Gaussian features into the StyleGAN architecture, informed by a comprehensive analysis of its structure. 

We validate the efficacy of our framework through both quantitative and qualitative experiments focused on self and cross-reenactment. Our key contributions are summarized as follows:

\begin{itemize}
    \item We present GaussianStyle, a framework that integrates 3D Gaussian splatting with StyleGAN representations. This integration enhances the controllability of head pose, facial expression, and fine-grained facial details, enabling high-quality volumetric avatar generation from monocular videos.

    \item We refine the hybrid triplane-Gaussian representation by introducing a temporal-aware design and an attention-based deformation module. This improves the deformability of Gaussian points, leading to more robust and accurate 4D face rendering.
    
    \item We design a pipeline that effectively maps dynamic 3D representations to the latent space of StyleGAN for volumetric 3D rendering. This approach requires training only a small number of parameters, achieving an easily editable neural representation with inference speeds exceeding 30 FPS while maintaining high fidelity.
    
\end{itemize}
\section{Related Work}
\label{sec:related_works}

\noindent \textbf{Video Portrait Animation.}
Mainstream approaches for facial reconstruction and animation primarily relied on 3D Morphable Models(3DMM)~\cite{deng2019accurate} or relying on implicit neural representations~\cite{zheng2022avatar, athar2022rignerf, gao2022reconstructing}. Neural Head portrait~\cite{grassal2022neural} employs neural networks to dynamically enhance the texture and geometry of head portraits, utilizing the FLAME model~\cite{FLAME:SiggraphAsia2017}. IMAvatar~\cite{zheng2022avatar} and INSTA~\cite{INSTA:CVPR2023} shifted towards using implicit geometry to overcome the limitations of mesh templates. The point-cloud-based models combine explicit point clouds with neural networks' implicit representations to enhance image quality~\cite{Zheng2023pointavatar}. Recent works~\cite{qian2023gaussianavatars, wang2024gaussianhead} have shifted the direction towards Gaussian Splatting for head modeling, aiming to leverage the benefits of rapid training and rendering while still achieving competitive levels of photorealism. GaussianAvatars~\cite{qian2023gaussianavatars} reconstructed head avatars through rigging 3D Gaussians on FLAME~\cite{FLAME:SiggraphAsia2017} mesh. MonoGaussianAvatar~\cite{chen2024monogaussianavatar} learned explicit head avatars by deforming 3D Gaussians from canonical space with Linear Blend Skinning (LBS) and simultaneously. GaussianHead~\cite{wang2024gaussianhead} adopted a motion deformation field to adapt to facial movements while preserving head geometry. FlashAvatar~\cite{xiang2024flashavatar} initializes Gaussian based on the UV coordinates and learns the deformation offset conditioned on tracking parameters. However, none of these methods considers the dynamic coordinate change of Gaussian points and thus cannot present a robust performance towards novel poses or camera views.

\noindent \textbf{StyleGAN-based Portrait Editing and Rendering} 
With diverse style distribution, StyleGAN significantly promotes the progress of facial editing~\cite{roich2022pivotal, song2021agilegan, yang2022Pastiche, gal2022stylegan}. DeformToon3D~\cite{zhang2023deformtoon3d} further extends geometry-aware 3D editing. Portrait rendering techniques also benefited from StyleGAN. StyleHEAT~\cite{yin2022styleheat} optimizes latent codes through inversion and leverages audio features for motion, further refined by OTAvatar~\cite{ma2023otavatar} that applies EG3D to perform geometry-aware rendering. Next3D~\cite{sun2023next3d} and IDE-3D~\cite{sun2022ide} disentangle semantics and geometry for 3D-aware controlled avatar rendering. However, these works mainly focus on aligned faces and are not applicable to avatars containing torso areas.
\begin{figure*}[t]
    \centering
    \includegraphics[width=\linewidth]{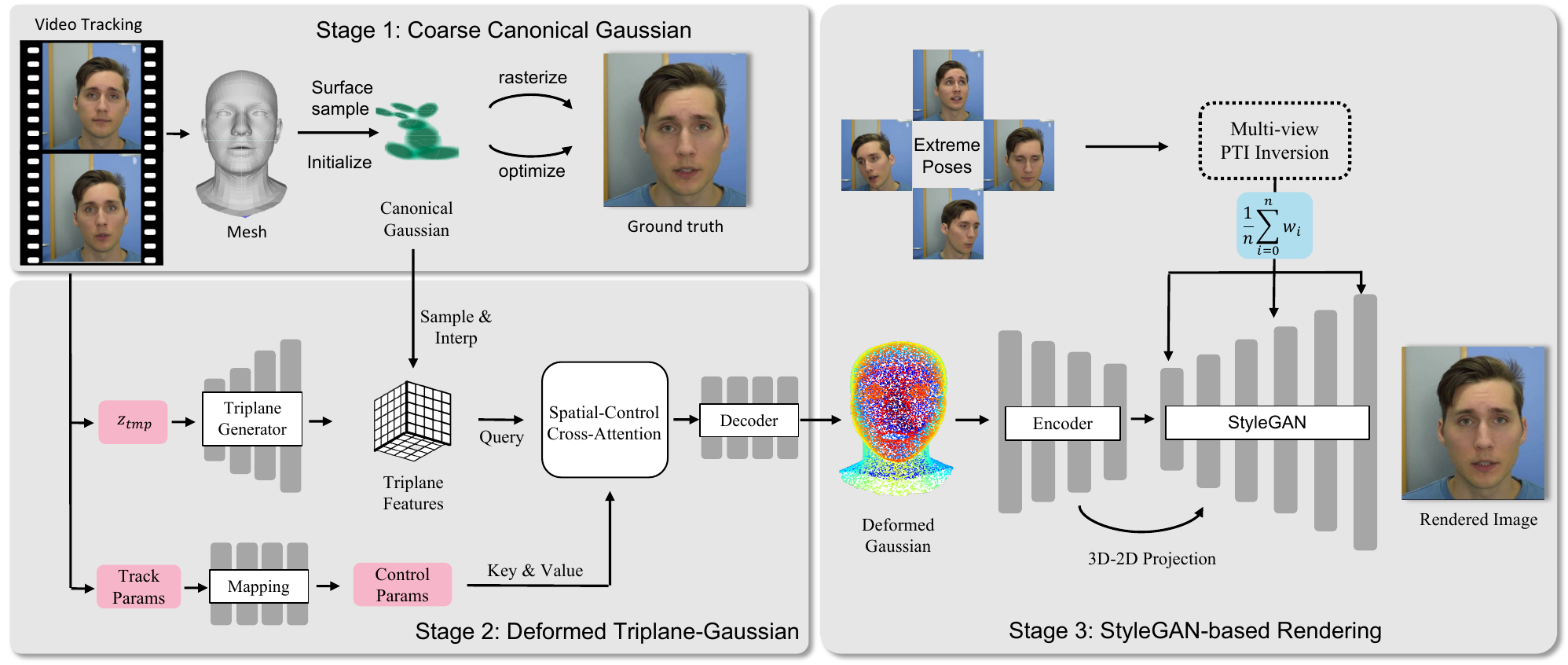}
    \vspace{-0.5cm}
    \caption{
        The proposed Tri-Stage training strategy includes StyleGAN-based Volumetric Rendering. In Stage 1, we construct static coarse canonical Gaussians. In Stage 2, Gaussians are queried from a temporal-aware triplane for attention-based deformation. In Stage 3, we initialize the StyleGAN through multi-view PTI initialization and project dynamic Gaussian prior into StyleGAN for volumetric rendering.
    } \label{fig:framework}
    \vspace{-0.2cm}
\end{figure*}

\section{Method}
As depicted in Fig.~\ref{fig:framework},  our framework combines Gaussian with StyleGAN~\cite{karras2019style} for Volume Rendering. StyleGAN's high-quality generation capability and style control proficiency make it suitable for our objectives. We first present a temporal-aware hybrid triplane-gaussian representation with attention-based deformation to achieve robust pose and expression control. (Sec.~\ref{sec:deformation}) To counteract the oversmooth problems, we explored an effective strategy of mapping Gaussian representation into StyleGAN's latent (Sec.~\ref{sec:mapping}).

\subsection{Deformable Triplane-Gaussian}
\label{sec:deformation}
\noindent \textbf{Temporal-ware Hybrid Representation}
Recent studies~\cite{zou2023triplane, wang2024gaussianhead} have demonstrated that hybrid triplane-Gaussian representations are effective in capturing continuous, structural, and low-dimensional features for 3D modeling. We extend this strategy to develop a deformable hybrid representation for 4D head modeling. Our approach employs a convolutional neural network generator, inspired by the StyleGAN architecture~\cite{karras2020analyzing}, to synthesize features within a triplane representation.
To incorporate temporal dynamics, we introduce a frame-specific latent code, denoted as $z_{tmp}$, into the generator for each input frame. For each 3D Gaussian centered at $\mu$, the coordinates are normalized, and corresponding features are obtained by interpolating the position on a regularly spaced 2D grid for each plane. These features are concatenated $\bigcup$ across dimensions to produce a final feature vector $F(\mu)$ or each canonical Gaussian position $\mu_c$:
\begin{equation}\label{eq:contraction}
    F(\mu) =\bigcup \mathrm{interp}\big(plane, \mathbf{P}(\mu)\big)
\end{equation}
where $\mathbf{P}(\mu)$ denotes a projection of $\mu$ onto the plane and `$\mathrm{interp}$' represents bilinear interpolation on the 2D grid. 

\begin{figure*}[t]
\centering

\includegraphics[width=\textwidth]{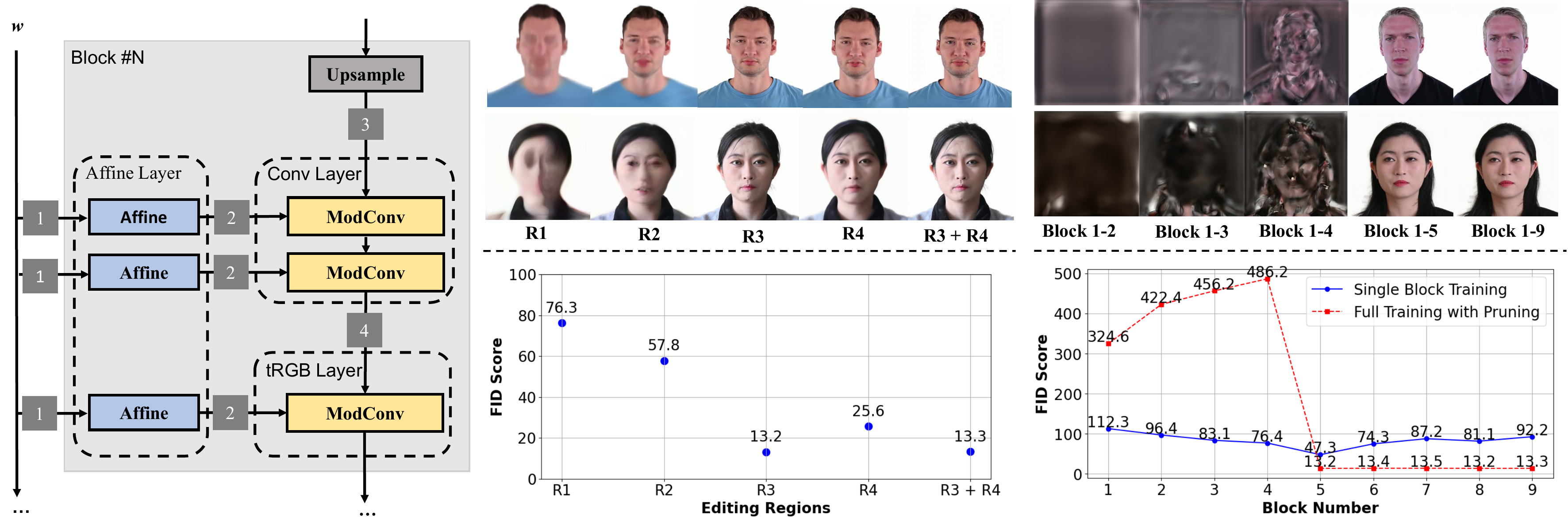}
   
\vspace{-0.2cm}
\caption{Left: Four regions within a single StyleGAN Block for features manipulation. Mid: Integration to R3 performs the best. R3+R4 does not bring improvement. Right: Blocks 1 to 5 are effective for volumetric projection. The upper refers to the block pruning results.}
\label{fig:combined}
\vspace{-0.2cm}
\end{figure*}

\noindent \textbf{Attention-based Deformation}
Traditional deformable Gaussian models~\cite{yang2023deformable3dgs, xiang2024flashavatar, Wu_2024_CVPR} typically concatenate conditioning parameters with Gaussian points to predict offsets for dynamic rendering. However, this approach overlooks a critical challenge inherent in dynamic 3D head modeling: the assumption that a fixed 3D coordinate in the canonical space will always correspond to the same facial region throughout the entire sequence.

To address this limitation and improve the correspondence between Gaussian points and conditioning parameters (such as facial expressions and head poses), we introduce a cross-attention mechanism. This mechanism fuses the spatial feature embeddings $F(\mu_c)$ of the canonical 3D Gaussians with the conditioning parameters, capturing how input expression and other factors influence the movement of the 3D Gaussians. The cross-attention mechanism layer $CA(\cdot)$ and MLP layer $FFN(\cdot)$, each connected via skip connections. The process is defined as follows:

\begin{equation}
\label{eq1}
F(\mu)' = \text{CA}(F(\mu), \mathbf{c}{n}) + F(\mu),
\end{equation}
\begin{equation}
\label{eq2}
Z(\mu) = \text{FFN}(F(\mu)') + F(\mu)',
\end{equation}
where the cross-attention is computed between the triplane feature $F(\mu)$ and the conditional feature $c_{n}$ of the $n$-th image frame. The output feature $Z(\mu) $ effectively integrates the conditioning information with the detailed facial features captured by each 3D Gaussian.

Finally, based on the condition-aware feature representation given the cross-attention, we leverage a deformation MLP $Deform(\cdot)$ for predicting the spatial dynamic offsets of Gaussians:
\begin{equation}
\Delta \mathbf{c}, \Delta \mathbf{\mu}, \Delta \mathbf{r}, \Delta \mathbf{s} = \text{Deform}(Z(\mu))
\label{eq:mlp}
\end{equation}

\subsection{Extended StyleGAN based on Gaussian}
\label{sec:mapping}

Pre-trained on the aligned FFHQ dataset, StyleGAN struggles with unaligned avatars commonly found in portrait videos. We defer a detailed analysis to the Appendix. Additionally, the lack of geometric awareness prevents StyleGAN from effectively handling novel view and pose reconstruction, which is essential for video avatar rendering. To address this limitation, we present a novel methodology that efficiently retains StyleGAN's pre-trained generalization abilities while encoding animatable 3D Gaussian representations into its latent space. This extension enhances its capability to generate and edit drivable video portraits. 

\noindent \textbf{Multi-view PTI initialization}
First, we employ PTI inversion~\cite{roich2022pivotal} from multiple images from the training dataset with extreme head poses to embed the target portrait within StyleGAN's latent distribution by subtly modifying the original model parameters. We first fix StyleGAN's parameters and optimize style code \( w \) to minimize the discrepancy between the generated and target images, indicating that \( w \) closely aligns with our target in the latent space. Subsequently, we fix \( w \) and fine-tune StyleGAN to enhance the similarity of the generated image to the target at this specific \( w \).

\noindent \textbf{Volumetric Projection to StyleGAN}
Next, we conduct a comprehensive analysis of StyleGAN's structure. Directly leveraging StyleGAN for unaligned 3D presentations is a non-trivial task. To resolve this issue, we investigate StyleGAN's architecture for the integration of dynamic volumetric representations from the Gaussians. Inspired by Pixel2Style2Pixel~\cite{richardson2021encoding}, we designed a Convolutional Encoder to encode Gaussian priors. However, directly formulating the 3D feature projection through manipulating style code \(w\) presents unwanted results. To explore the effective strategy of volumetric feature projection to StyleGAN, we further investigate StyleGAN's architecture. Fig.\ref{fig:combined} Left presents the four regions for 3D projection within a single StyleGAN Block. R1 refers to style code \(w\) manipulation. We in addition propose R2: Altering the style latent post Affine-Layer mapping. R3: Integrate the prior volumetric feature with the Conv-Layer feature. R4: Integrate the prior volumetric feature with the tRGB-Layer feature. We defer the details of the model structure in the Appendix.

\noindent \textbf{Volumetric Feature Integration}
We train StyleGAN with modification within four regions over 10 epochs. Our analysis, illustrated in Fig.~\ref{fig:combined} Middle, reveals that R1, R2, and R4 yield imprecise facial details, albeit preserving the general facial positioning within images. This suggests that modifications at the latent code, Affine-Layer levels, or intermediate tRGBs fail to impart adequate texture detail to StyleGAN. Conversely, we discover that feature integration to Conv-Layers suffices for embedding volumetric Gaussian priors into StyleGAN. We in addition explored R3 + R4, resulting in no performance difference. Consequently, we opt for exclusively modifying only the Conv-Layers.

\noindent \textbf{Effective StyleGAN Blocks} 
Our subsequent investigation focuses on determining the effective blocks for the modification. We adopt two strategies for the investigation. (1) Integrate projection to a single block (2) Integrate projections across all StyleGAN blocks during training, and prune projections during inference. Shown in Fig.~\ref{fig:combined} Right, our study for (1) reveals that Blocks 1 to 5 (4x4 to 64x64) are effective with Block 5 significantly better than others. In addition, (2) presents that during pruning, Blocks 1 to 4 are instrumental for geometry while Block 5 refines texture. The others can be pruned with no detrimental impact. We thus choose the first five blocks for the volumetric feature projection.

\subsection{Training strategy of Volumetric Rendering}
\noindent \textbf{Canonical Gaussian}
We first reconstruct the mean shape of the talking face, by optimizing the positions of 3D Gaussians and the triplane generator. We initialize the 3D Gaussian center positions by sampling the surface of the mesh from video tracking. This preserves the shape topology of the face and landmarks of the target avatar. 
Unlike conventional Gaussian rendering that only considers RGB, we render a 32-channel volumetric feature for better 3D representation. We employ both $L_1$ loss and LPIPS loss for aligning synthesized images with ground truth. We focus on the first three channels of the Gaussian volumetric representation $I_{gs}^{(1:3)}$, comparing them against RGB ground-truth images $I_{gt}$. Other channels will be deferred to StyleGAN-based rendering.

\begin{equation}
\mathcal{L}_{rgb} = || I_{gs}^{(1:3)} -  I_{gt}||_{1}
\label{loss_stage_1}
\end{equation}

\noindent \textbf{Deformation}
We optimize the triplane generator, cross-attention, and deformer for deformation. In addition to the previous L1 loss, we additionally focus on the eye and mouth regions for learning the expression. 

\begin{equation}
\mathcal{L}_{lmk} = || R_n(I_{gs}) -  R_n(I_{gt}) ||_{1}
\label{loss_stage_2}
\end{equation}
Rn is either the eyes or mouth region extracted using RoI-align on the bounding boxes calculated using landmarks.

\noindent \textbf{StyleGAN-based Volumetric Synthesis}
The final stage optimizes the triplane generator, the deformer, the encoder, and the projection layers while freezing the whole StyleGAN. We employ the LPIPS loss~\cite{johnson2016perceptual} between synthesized and ground-truth images.
\begin{equation}
\vspace{-0.1cm}
\mathcal{L}_{perp} = \text{LPIPS}(I_{gan}, I_{gt}).
\label{loss_perp}
\end{equation}
To further enhance image fidelity, we introduce a conditional discriminator, using UV maps as a condition to compare generated images with ground truth. This method, employing conditional adversarial loss~\cite{mirza2014conditional}:

\begin{equation}
\begin{aligned}
\begin{aligned}
\mathcal{L}_{cGAN}\left(G, D\right)= & \mathbb{E}_{I_{gt}, uv}\left[\log D\left(I_{gt}, uv\right)\right]\\
& + \mathbb{E}_{uv}\left[\log \left(1-D\left(uv, I_{gan}\right)\right]\right.
\end{aligned}
\label{loss_adv}
\end{aligned}
\end{equation}
where $D$ aims to distinguish between $\left\{\left(I_{gt}, uv\right)\right\}$ and $\left\{\left(I_{vr}, uv\right)\right\}$, and $(\cdot, \cdot)$ denotes concatenation.

\begin{figure*}[t]
    \centering
    \includegraphics[width=0.95\linewidth]{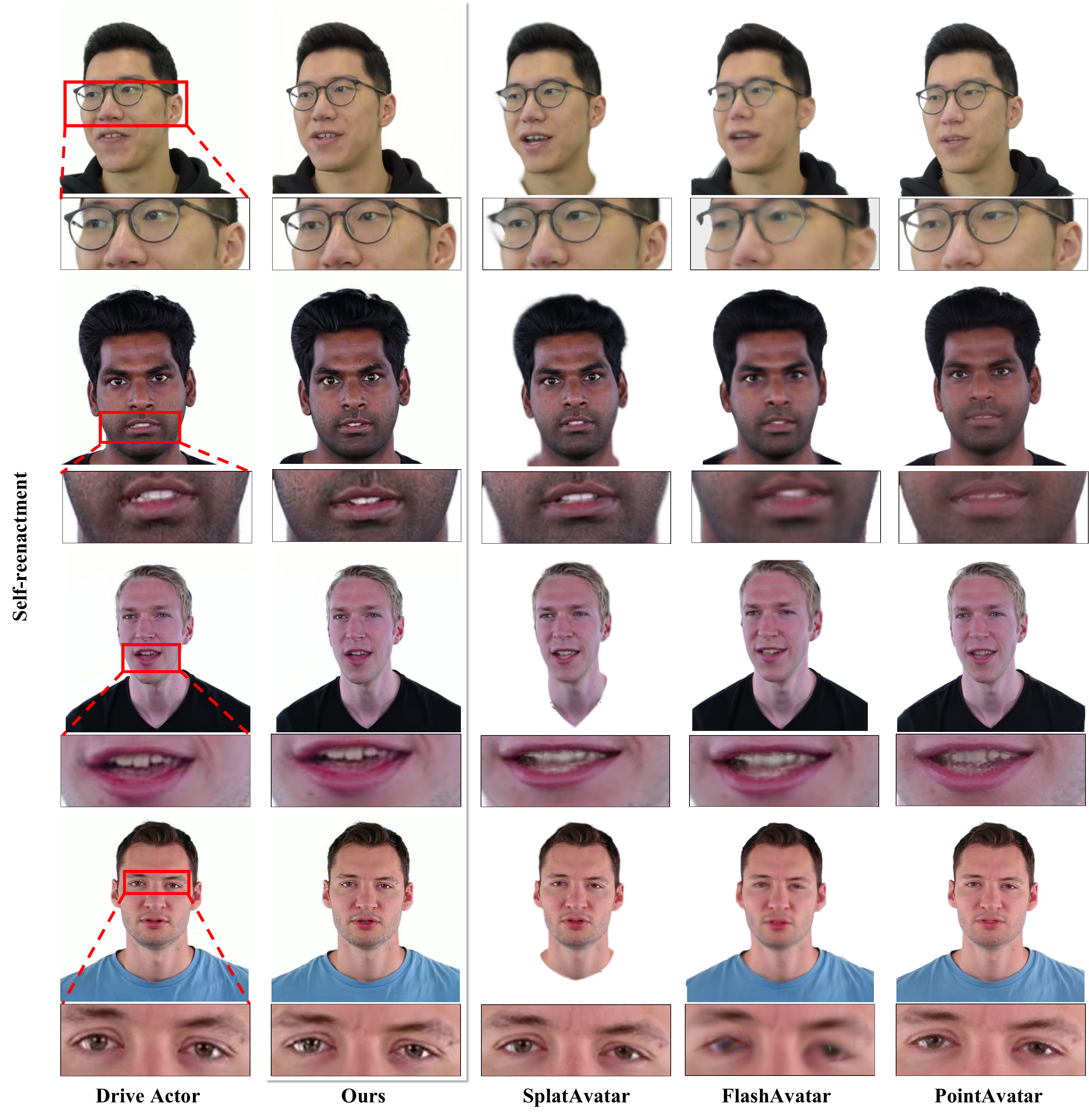}
    \vspace{-0.5cm}
    \caption{ Our model outperforms other monocular avatar rendering methods in detail such as eyes and teeth.
    } \label{fig:self-reenactment}
    \vspace{-0.3cm}
\end{figure*}

\begin{figure*}[t]
    \centering
    \includegraphics[width=0.95\linewidth]{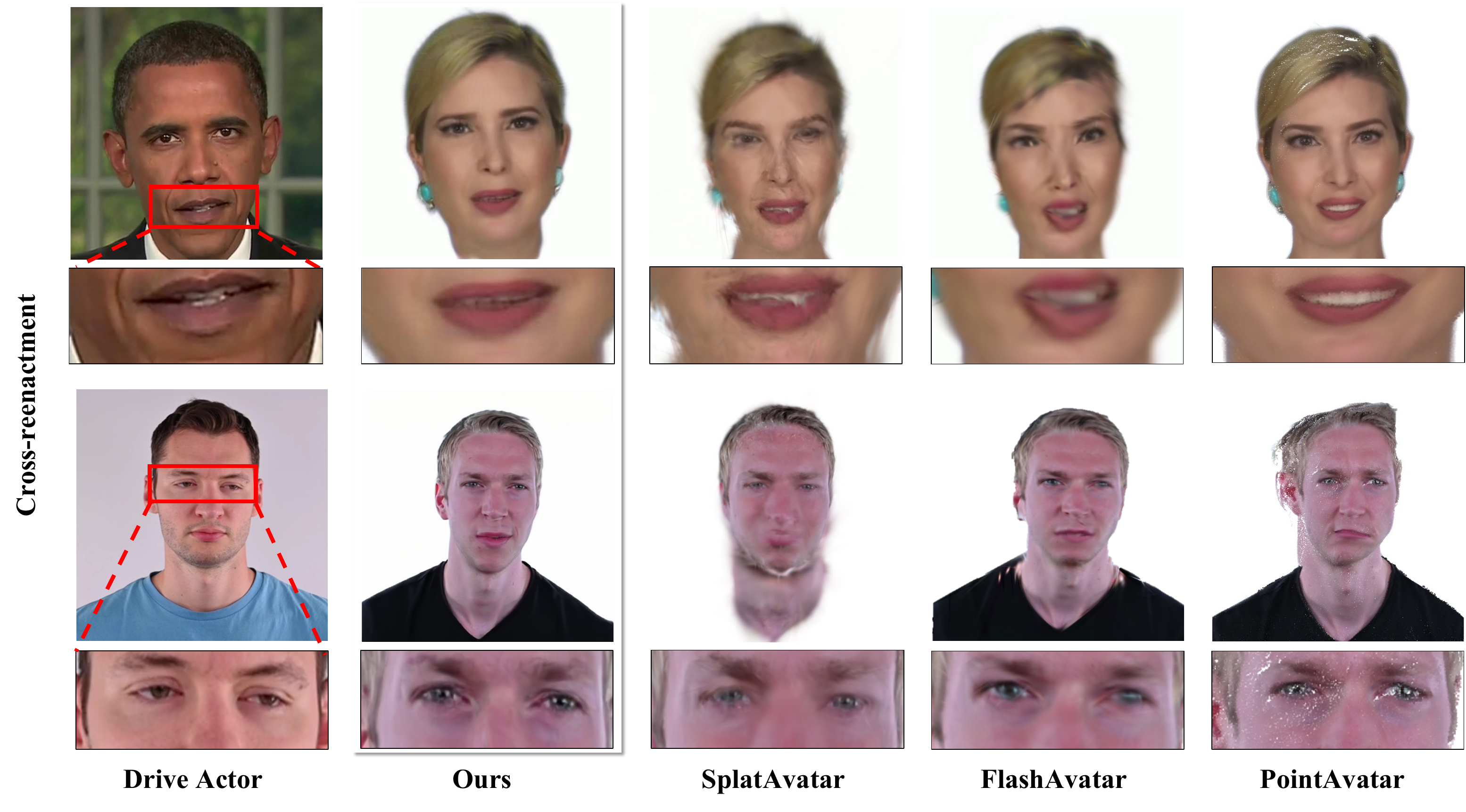}
    \vspace{-0.5cm}
    \caption{Other methods are not robust to novel views, expressions, or head poses and thus exhibit noisy point clouds and blurred results.
    } \label{fig:cross-reenactment}
    \vspace{-0.1cm}
\end{figure*}

\section{Experiment}

\paragraph{Implementation Details}
We implement our model with PyTorch and a single A6000 GPU. We use StyleGAN2 distill-version, MobileStyleGAN 1024x1024, pre-trained on FFHQ as the generator for all studies. For Coarse Gaussian and Deformation jointly, We train the model for 10,000 iterations. For the StyleGAN-based synthesis stage, we train the triplane generator, the deformer, the encoder, and the projection layers while freezing the StyleGAN with a batch size of 4 for 50,000 iterations. The Adam optimizer~\cite{kingma2017adam} is adopted for all learnable parameters with a learning rate of $1e^{-4}$. We present the details of the encoder and projection layer, with the number of trainable parameters in the Appendix.

\paragraph{Dataset}
Our method takes a monocular video as input and leverages expression parameters from tracking to achieve video portrait rendering and editing. We primarily conduct experiments on data from INSTA~\cite{muller2022instant}, NeRFFace~\cite{Gafni_2021_CVPR}, and NerfBlendShape~\cite{Gao2022nerfblendshape} and self-recorded high-resolution videos. Each data sample captures a diverse range of facial motions in an average of 5-minute duration. All videos are resized to 1024$^2$ for our model. We divide the training set and testing set from each video to 80\% and 20\% of all frames, respectively. We conduct experiments comparing self/cross-reenactment with the current drivable portrait rendering techniques for avatar reconstruction and control.

\paragraph{Baseline Methods}
We benchmark GaussianStyle against the following methods for monocular video avatar rendering: (1) FlashAvatar~\cite{xiang2024flashavatar}, (2) PointAvatar~\cite{Zheng2023pointavatar}, (3) SplatAvatar~\cite{SplattingAvatar:CVPR2024}. We do not include GaussianAvatars~\cite{qian2023gaussianavatars} and Gaussian-Head-Avatar~\cite{xu2023gaussianheadavatar}, which reconstruct human heads from multi-views, unlike single-view monocular videos. We defer the comparison with NeRF-based and StyleGAN-based rendering methods in the Appendix.

\subsection{Quantitative Evaluation}
\paragraph{Evaluation Metrics} We evaluate the effectiveness of our method on three aspects: (1) F-LMD~\cite{chen2018lip}: The differences in head pose and facial expression landmark positions calculated via MediaPipe~\cite{lugaresi2019mediapipe}. (2) The Sharpness Difference (SD)~\cite{mathieu2015deep}: It is used to evaluate the sharpness difference between the source and generated images at the pixel level. (3) Image Spatial Quality: we adopt the PSNR, the Learned Perceptual Image Patch Similarity (LPIPS)~\cite{zhang2018unreasonable}, and SSIM for image generation quality evaluation.

\noindent \textbf{Evaluation Results} Table~\ref{table_1} summarizes the quantitative results, where our method consistently outperforms the baselines in terms of image quality (PSNR, LPIPS), sharpness (SD), and motion accuracy (F-LMD). The significant improvement in LPIPS and SD highlights our method's ability to enhance detail intensity and perceptual similarity.

\begin{table}[t]
\footnotesize
\setlength{\tabcolsep}{0.8mm} 
\centering
\vspace{-.1cm}
\begin{tabular}{cccccccc}
\hlinew{1.15pt}
\multirow{3}{*}{Methods} &F-LMD$\downarrow$ &SD$\downarrow$ &PSNR$\uparrow$&LPIPS$\downarrow$ &MOS$_1$ &MOS$_2$ &MOS$_3$\\
\cline{2-8}
&\multicolumn{4}{c|}{Quantitative Results} &\multicolumn{3}{c}{User Study}\\
\cline{2-8}
&\multicolumn{4}{c|}{Dataset A}&\multicolumn{3}{c}{Self-Reenactment}\\
\hline
FlashAvatar & 2.96 & 9.43  & 27.97  & 29.42 & 2.26 & 1.97  & 2.05    \\
PointAvatar & 2.55 & 8.42 & 28.39  &  23.64 & 3.67 & 4.03 & 3.96     \\
SplatAvatar & 2.88 & 4.54 & 32.53  & 25.47  & 3.88 & 3.76 & 4.21  \\
\rowcolor{mygray} Ours &\textbf{2.42} &\textbf{3.38} &\textbf{34.43} & \textbf{13.14}& \textbf{4.47} & \textbf{4.27} & \textbf{4.73} \\
\hline
&\multicolumn{4}{c|}{Dataset B}& \multicolumn{3}{c}{Cross-Reenactment}  \\
\hline
FlashAvatar & 3.82 & 10.67  & 25.43  & 25.35 & 1.51 & 2.12  & 1.13  \\
PointAvatar &2.64 & 8.24 & 26.19 &  21.42 & 3.31 & 2.89 & 3.80  \\
SplatAvatar & 3.11 & 5.23 & 28.32   & 19.46 &3.45 & 2.76 & 3.11  \\
\rowcolor{mygray} Ours &\textbf{2.31} &\textbf{2.84}   & \textbf{30.44} &\textbf{11.82} &\textbf{4.21} &\textbf{3.83} & \textbf{3.89}  \\
\hlinew{1.15pt}
\end{tabular}
\vspace{-.2cm}
\caption{(1) Left: Quantitative results of FlashAvatar~\cite{xiang2024flashavatar}, PointAvatar~\cite{Zheng2023pointavatar}, SplatAvatar~\cite{SplattingAvatar:CVPR2024}. We bold the best. The values of SD and LPIPS are multiplied by $10^{-1}$ and $10^{2}$ respectively. (2) Right: The MOS score for human evaluation. Each one comes from a 5-point Likert scale (\textit{1-Bad, 2-Poor, 3-Fair, 4-Good, 5-Excellent}). The closer to $5$ the better, we bold the best.}
\label{table_1}
\vspace{-0.2cm}
\end{table}

\subsection{Qualitative Results} 
\label{Qualitative Results}

\begin{figure*}[t]
    \centering
    \includegraphics[width=1\linewidth]{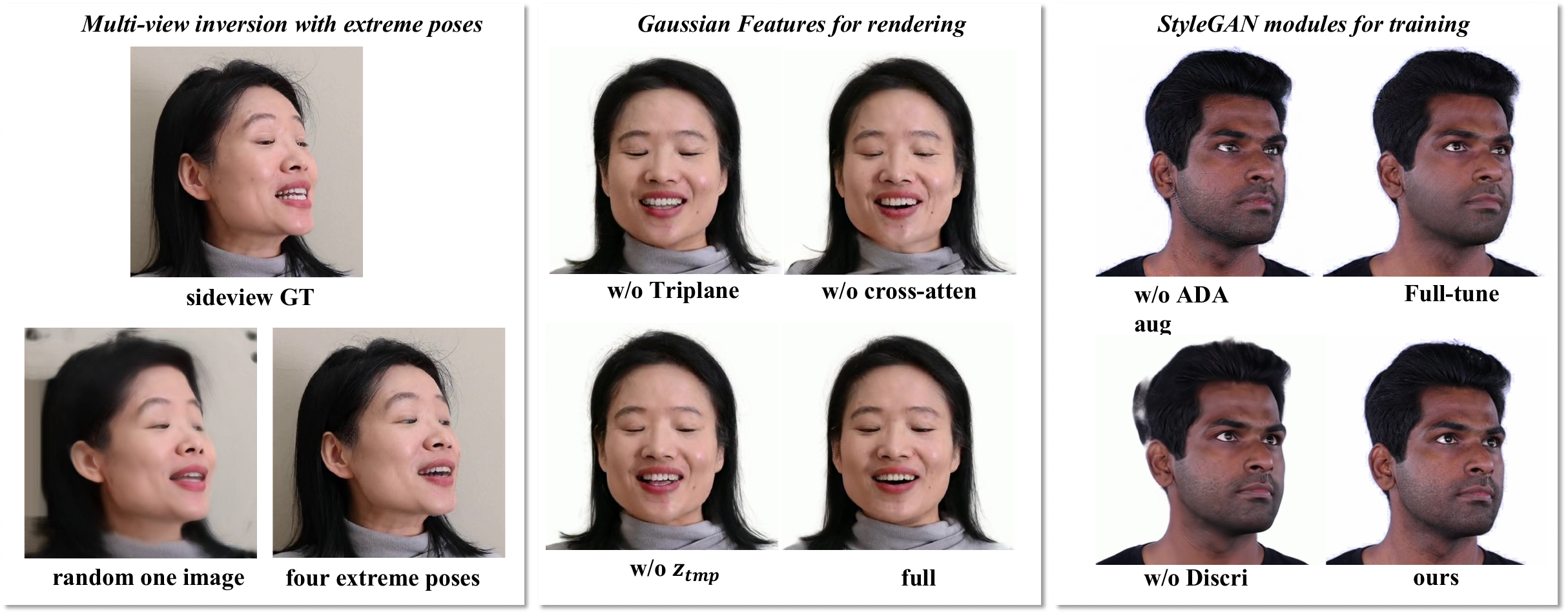}
    \vspace{-0.5cm}
    \caption{Ablations for multi-view PTI inversion, gaussian-feature contribution, and StyleGAN training.
    } \label{fig:ablation}
    \vspace{-0.2cm}
\end{figure*}

\begin{table*}[t]
  
  \label{tab:ab}
  \renewcommand{\tabcolsep}{1.8pt}
  \small
\begin{subtable}[!t]{0.3\linewidth}
    \centering
  \begin{tabular}{cccc}
  
    \toprule
    {\it View Init.}     & LPIPS\(\downarrow\)      &PSNR\(\uparrow\)  &SSIM\(\uparrow\)\\
    \midrule
    1 random & 18.51 &32.47 &0.842 \\
    4 random & 16.42 &33.36 &0.839 \\
    2 extreme & 15.63 &34.02 & 0.847 \\
    8 extreme & 14.57 & 34.21 & 0.873 \\
    \rowcolor{mygray}Ours & \textbf{13.14} & \textbf{34.43} & \textbf{0.886}\\
    \bottomrule
  \end{tabular}
  \caption{The effect of multi-view PTI Inversion}
  \label{tab:ab_repr}
\end{subtable}
\hspace{\fill}
\begin{subtable}[!t]{0.31\linewidth}
    \centering
  \begin{tabular}{cccc}
    \toprule
    {\it Gauss Feats.}  & PSNR\(\uparrow\)  &SSIM\(\uparrow\)  &Train\(\downarrow\)\\
    \midrule
    w/o triplane & 30.32 & 0.818 & \textbf{15 min} \\
    w/o cross-atten & 32.38 & 0.834 & \textbf{15 min}\\
    w/o $z_{tmp}$ & 33.15 & 0.857 & \textbf{15 min}\\
    w/o init & 28.14 & 0.764 & 25 min\\
    \rowcolor{mygray} Ours & \textbf{34.43} & \textbf{0.886} & \textbf{15 min}\\
    \bottomrule
  \end{tabular}
  \caption{The effect of different types of features.}
  \label{tab:ab_face}
\end{subtable}
\hspace{\fill}
\begin{subtable}[!t]{0.31\linewidth}
    \centering
  \begin{tabular}{ccccc}
    \toprule
    {\it StyleGAN feats.} & PSNR\(\uparrow\)  &Train\(\downarrow\) & In-Speed\(\uparrow\)\\
    \midrule
    StyleGAN2 & \textbf{35.53} & 3.5h & 18fps \\
    w/o ADA-aug & 30.62 & 2.5h & \textbf{30fps} \\
    w/o Discri & 30.44 & \textbf{2h} & \textbf{30fps} \\
    full-tune & 34.32 & 5.5h & \textbf{30fps} \\
    \rowcolor{mygray} MobileStyleGAN & 34.43 & 2.5h & \textbf{30fps}\\
    \bottomrule
  \end{tabular}
  \caption{The effect of StyleGAN training.}
  \label{tab:ab_hand}
\end{subtable}
\vspace{-1mm}
\caption{Ablations of our method. We vary views for initialization, feature types, and StyleGAN training to investigate their effectiveness.}
\end{table*}

\begin{figure*}[t]
    \centering
    \includegraphics[width=1\linewidth]{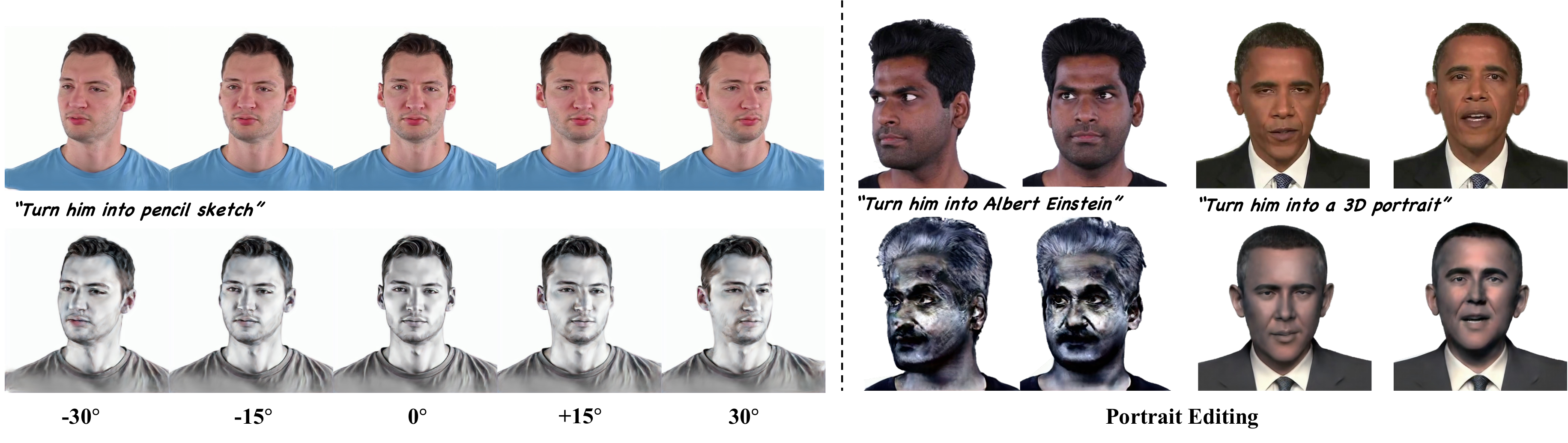}
    \vspace{-0.8cm}
    \caption{Left: We apply -30 to +30 degree rotation to the camera. The visualized results are consistent over multi-views. Right: 200 images are sampled for fine-tuning. IN2N is applied for text-driven editing guidance and takes about 10 minutes for each subject.
    } \label{fig:application}
    \vspace{-0.2cm}
\end{figure*}

Fig.\ref{fig:self-reenactment} and Fig.\ref{fig:cross-reenactment} provide qualitative comparisons for self-reenactment and cross-reenactment scenarios, respectively. The highlighted zoomed-in regions of each generated image demonstrate the distinct strengths and weaknesses of each method. FlashAvatar produces noisy point clouds and blurred outputs due to inadequate regularization of implicit pose and expression deformations. SplatAvatar excels in shape reconstruction but struggles with appearance recovery, particularly in challenging areas like the torso, mouth, and eyes during cross-reenactment. Point-Avatar manages to recover specific features such as glasses but delivers overly smooth facial representations, failing to capture subtle expressions and clear teeth despite significant computational demands. In cross-reenactment, the baseline methods generally exhibit blurring and noise due to out-of-domain expressions and head poses. In contrast, our approach, leveraging cross-attention for deformation, maintains a robust representation of 3D points conditioned on tracking parameters. This enables precise recovery of high-frequency details and accurate control of head pose and facial expressions, outperforming all comparative baselines across both self-reenactment and cross-reenactment scenarios.

\noindent \textbf{User Study} We conducted a user study to evaluate the visual quality of our method, following the protocols established in Deep Video Portrait~\cite{kim2018deep}. The study included 40 videos—20 for self-reenactment and 20 for cross-reenactment. We recruited 20 participants via Amazon Web Services (AWS) to assess the quality based on several criteria, using the Mean Opinion Scores (MOS) rating system. Participants rated the videos on: (1) MOS$_1$: \textit{“How is the image quality in the video?”}, (2) MOS$_2$: \textit{“How realistic does the video appear?”}, and (3) MOS$_3$: \textit{“Are the facial motions synchronized between the two videos?”}. The videos were presented in random order to capture participants’ initial impressions. As shown in the right section of Table~\ref{table_1}, our method outperformed others across all criteria, demonstrating superior video quality, realism, and motion synchronization.

\subsection{Ablation Studies}


\noindent \textbf{Multi-view PTI Inversion}
We investigate strategies of PTI inversion for StyleGAN initialization: (1) a single random image, (2) multiple random images, and (3) multiple images depicting extreme poses. In Fig.~\ref{fig:ablation}, PTI with a single image produces blurred results for side views. However, initializing with four images capturing extreme poses along the x and y axes effectively addresses this issue, shown in Tab~\ref{tab:ab_repr}. Inversion with additional images lowers the performance.

\noindent \textbf{Rendering Features}
To assess the contribution of each component, we design the following variations (1) w/o triplane: Pure Gaussian representation without triplane.  (2) w/o cross-atten: No cross-attention but only MLP for deformation used in DeformableGaussian~\cite{yang2023deformable3dgs} and FlashAvatar~\cite{xiang2024flashavatar} for Gaussian offsets. (3) w/o triplane generator: triplane not generated by a Stylegan-like generator but learned as in GaussianHead~\cite{wang2024gaussianhead} and 4DGaussian~\cite{Wu_2024_CVPR} (4) w/o init: No Gaussian initialization based on mesh but optimized from scratch. As detailed in Tab.~\ref{tab:ab_face} and Fig.~\ref{fig:ablation}, our design of Deformable Hybrid Triplane-Gaussian representation significantly improves the rendering quality.

\noindent \textbf{StyleGAN Features}
To assess the contribution of each component, we design the following variations (1) StyleGAN: all other modules remain the same except replacing MobileStyleGAN with StyleGAN2 (2) w/o ADA-aug: no data augmentation used in StyleGAN-ADA~\cite{Karras2020ada} applied during training (3) w/o Discri: No distriminator used during training. (4) full-tune: No PTI inversion and fully fine-tuned StyleGAN without freezing. As detailed in Tab.~\ref{tab:ab_face} and Fig.~\ref{fig:ablation}, our strategy that preserves StyleGAN's latent distribution helps reduce the training time while improving quality. In addition, the data augmentation significantly improves the performance in the monocular video setting. MobileStyleGAN achieves a compatible performance but much faster than StyleGAN2. We thus take MobileStyleGAN as the final generator.

\subsection{Applications: 3D Editing and Novel View}

GaussainStyle is a general representation of volumetric head avatars and is not restricted to real monocular video reconstruction but can be easily extended for novel view synthesis and portrait editing.

\noindent \textbf{Novel View Synthesis}
We fix the control parameters and apply -30 to +30 degree rotation to the camera. Fig.~\ref{fig:application} shows our model is robust in novel view synthesis.

\noindent \textbf{Portrait Editing}
Following Instruct-NeRF2NeRF~\cite{haque2023instruct} (IN2N), we use InstructionPixel2Pixel~\cite{brooks2022instructpix2pix} for guidance. After training on realistic faces, we randomly select 200 images from the dataset for iterative dataset update. We freeze the Gaussian, deformation modules, and StyleGAN, with only the projection layer trainable. All other settings remain the same as in IN2N. The typical editing time is around 10 minutes per subject on a single GPU.

\vspace{-0.1cm}
\section{Conclusion}
\label{Conclusion}
\vspace{-0.1cm}

In this work, we present GaussianStyle, a novel framework that integrates 3D Gaussian splatting with StyleGAN to achieve high-fidelity volumetric avatar generation. Our temporal-aware tri-plane, combined with an attention-based deformation module, refines the Gaussian representation, resulting in more robust and accurate dynamic face rendering. Additionally, our pipeline for mapping dynamic 3D representations to the latent space of StyleGAN maintains the generalization capabilities of the pre-trained model while enabling editable neural representations. We achieve an inference speed of over 30 FPS while maintaining high fidelity across novel-view and self/cross-reenact synthesis scenarios.
{
    \small
    \bibliographystyle{ieeenat_fullname}
    \bibliography{main}
}

\end{document}